\documentclass[10pt,twocolumn,letterpaper]{article}

\usepackage{blindtext}
\usepackage{graphicx}
\usepackage{color}
\usepackage{lineno}
\usepackage{multirow}
\usepackage{fancyhdr}
\usepackage{amsmath}
\usepackage{amsthm}
\usepackage{amssymb}
\usepackage{textcomp}
\usepackage{mdwmath}
\usepackage[tight,footnotesize]{subfigure}
\usepackage{fancyhdr}
\usepackage[utf8]{inputenc}
\usepackage{pifont}
\usepackage{bm}
\usepackage{soul}
\usepackage{amsfonts}
\usepackage[utf8]{inputenc}
\usepackage{setspace}
\usepackage{soul,xcolor}
\usepackage{wrapfig}
\usepackage{gensymb}

\usepackage{times}
\usepackage{epsfig}
\usepackage{array}
\usepackage{placeins}

\begin{document}
\title{From 2D to 3D Geodesic-based Garment Matching}
\author{\parbox{16cm}{\centering
    {\large Meysam Madadi$^{1,2}$, Egils Avots$^{5}$\footnote{Meysam Madadi and Egils Avots collaborated equally.}, Sergio Escalera$^{1,3}$, Jordi Gonz\`alez$^{1,2}$, Xavier Bar\'o$^{1,4}$, Gholamreza Anbarjafari$^{5,6}$}\\
    {\normalsize
    $^1$ Computer Vision Center, Edifici O, Campus UAB, 08193 Bellaterra (Barcelona), Catalonia Spain\\
    $^2$ Dept. of Computer Science, Univ. Aut\`onoma de Barcelona (UAB), 08193 Bellaterra, Catalonia Spain\\
    $^3$ Dept. Mathematics and Informatics, Universitat de Barcelona, Catalonia, Spain\\
    $^4$ Universitat Oberta de Catalunya, Catalonia, Spain\\
    $^5$ iCV Lab, Institute of Technology University of Tartu, Tartu, Estonia\\
    $^6$ Department of Electrical and Electronic Eng., Faculty of Engineering, Hasan kalyoncu University, Gaziantep, Turkey}}
}
\date{\vspace{-5ex}}
\maketitle

\begin{abstract}
\looseness=-1 A new approach for 2D to 3D garment retexturing is proposed based on Gaussian mixture models and thin plate splines (TPS). 
An automatically segmented garment of an individual is matched to a new source garment and rendered, resulting in augmented images in which the target garment has been retextured by using the texture of the source garment. We divide the problem into garment boundary matching based on Gaussian mixture models and then interpolate inner points using surface topology extracted through geodesic paths, which leads to a more realistic result than standard approaches. We evaluated and compared our system quantitatively by mean square error (MSE) and qualitatively using the mean opinion score (MOS), showing the benefits of the proposed methodology on our gathered dataset.
\end{abstract}

\section{Introduction}

As shopping for garments is increasingly moving to a digital domain, the next step after just seeing the desired clothes is to virtually try them on. The focus of this paper is an application for garment retexturing where the images of the person are captured with a Kinect-2 RGB-D camera. There are several steps between taking an RGB-D picture and displaying the final result with a retextured garment. These steps involve segmentation of the garment, garment matching and surface retexturing. The novelty of this paper lies in the retexturing part, which involves several challenges. First, a coordinate map must be created between the image of the new texture and the image that is being retextured. This problem is especially difficult in the case of non-rigid and easily transformable surfaces like clothes. 
Another challenge is to shade the new texture correctly. It is possible to use the colour information of the original image, but the lighting, intensity and the original colour of the surface are usually not previously known and must be estimated.

There exist several standard methods for projecting textured surfaces on screen. The simplest shading methods work only by using surface normals independently without considering the overall surface, attempting to estimate the brightness of the surface given some known viewer and light source direction. Examples of this kind of method
are the Gouraud shading, Phong shading, and Blinn-Phong shading~\cite{shirley2009fundamentals}. However, these methods do not support shadows.

\begin{figure*}[!ht]
  \centering
  \includegraphics[width=0.9\textwidth]{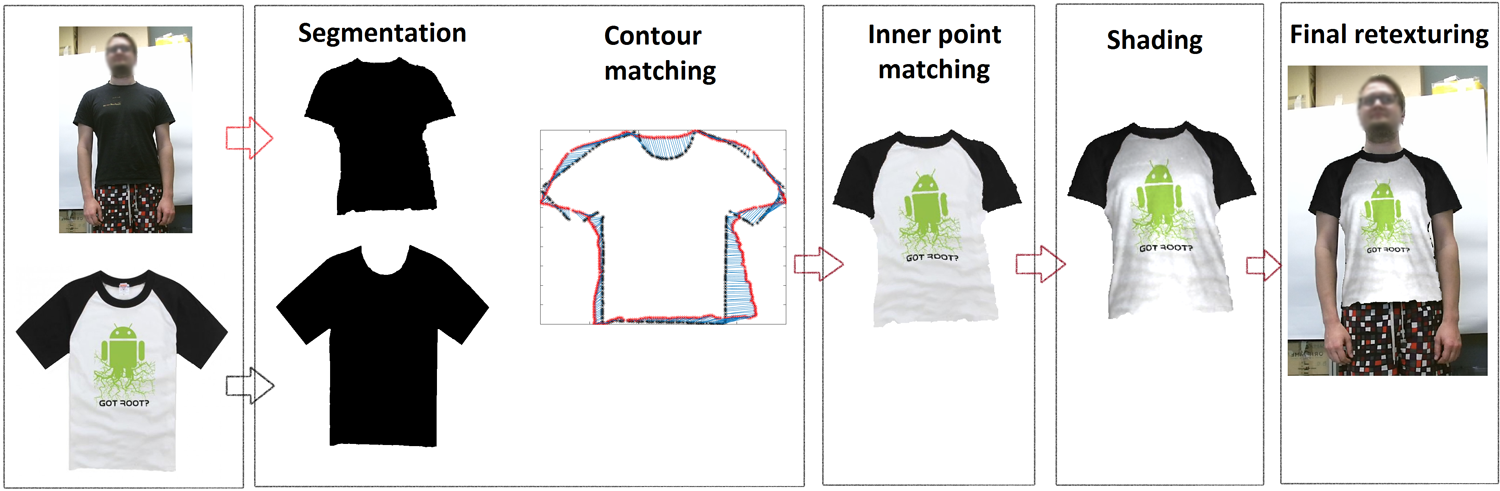}
  \caption{Overview of the proposed retexturing method.}
  \label{fig:flowChart}
\end{figure*}

The proposed automatic retexturing method, after the segmentation stage uses point set registration method~\cite{bing2010pointsetreg} to find correspondence between the outer 2D contours of the person and the target garment. After the contour matching, the surface topology of the flat 2D garment is approximated using geodesic distance in a global closed form solution using thin plate spline (TPS) \cite{bookstein1989} and the final result is superimposed onto the segmented area.

The rest of the paper is organised as follows: Section 2 discusses related work in the field of object retextureing, Section 3 gives a detailed description of the proposed retexturing method and Section 4 shows and compares the results obtained by using two different mapping methods. Finally, Section 5 concludes the paper.

\section{Literature review}
Due to the fact that an actual try-on of clothes is time-consuming, a virtual alternative has always been desired, and many researchers have been engaged in developing novel strategies and systems to perform such a task~\cite{tong2012scanning,zhou2012image,hauswiesner2011free,daneshmand2015real,sengupta2013virtual}. It requires scanning, classification of the body based on gender and size, 3D modelling~\cite{zhang2009method,harvent2013multi,fezza2015color} and visualisation.
Constrained texture mapping and parametrisation of triangular mesh are some popular examples, although they suffer from some deficiencies such as finding the parameter values and manual adjustments~\cite{ma2015foldover,liu2008local}. Many researchers have also suggested methodologies for visually fitting garments onto the human body based on dense point clouds~\cite{henry2010rgb,barone2012three}. 

The matching problem stage can be defined as a correspondence problem, which incorporates pair-wise constraints.
Hence, it is often solved with a graph matching approach~\cite{Henry2012,Zhou2012,fan2015object}, which is especially suitable for
deformable object matching. Furthermore, additional constraints can be added to the framework in order to reduce the computation
time (e.g. clearly, each cloth type is constrained to the body part where it is dressed), or in order to take problem-specific aspects
into account.

There exist various techniques for conducting a mapping from 2D image texture space to a 3D surface. Some examples are intermediate 3D shape~\cite{Bier1986}, direct drawing onto the object~\cite{Hanrahan1990}, or using an exponential fast marching method by applying geodesic distance~\cite{Xu2014,traumann2015accurate}. Many researchers have devoted special attention trying to enhance the realism of virtual garment representation during the last decade~\cite{Chang2014}. 
One of the most frequently used texture fitting methods was proposed by Turquin et al.~\cite{Turquin2007},
which allows the users to sketch garment contours directly onto a 2D view of a mannequin. The initial algorithm has been
further enhanced by many other researchers~\cite{Yaseen2013,Zhang2013}.

Another popular way of mapping a 2D texture onto a 3D surface is by using a single image~\cite{Zhou2013}. As proposed by~\cite{Zhou2012},
an estimation of a 3D pose and shape of the mannequin is followed by constructing an oriented facet for each bone of a mannequin
according to angles of the pose, and projecting the 2D garment outlines into corresponding facets. Eckstein et al.~\cite{Eckstein2001}
proposed a constrained texture mapping algorithm, which can be used for 2D and 3D modelling, and multi-resolution texture mapping and
texture deformation, but it may produce a Steiner vertex effect when a simple solution does not exist. Kraevoy et al.~\cite{Kraevoy2003} introduced a method based on iterative optimisation of a constrained texture mapping method.
In their method, it is a requirement to specify the corresponding constraint points on the grid model and texture image,
the parametrised mesh. Later, Yanwen et al.~\cite{Yanwen2005} reported a constrained texture mapping method
based on harmonic mapping, with interactive constraint selection by the user; the method produces high efficiency, real-time optimisation, and adjustment of mapping results. The  
block based
constrained texture mapping methods are also used in order to bring higher speed and lower computational costs~\cite{lui2013texture}.

\section{Retexturing approach}
In this paper, we propose a new automatic retexturing method covering the stages of segmentation, 2D to 3D garment matching and rendering. We use a Kinect~2 device to capture scene information. As preprocessing, we use RGB, depth and infrared images of the Kinect and segment out the garment from the background. The segmented depth image is used to compute retexturing from a source 2D flat garment image. We reduce the problem of surface point matching to an interpolating problem by using garment contour matching. The interpolation process takes surface topology into account using geodesic distance in a global closed form solution using thin plate spline (TPS) \cite{bookstein1989}. Thus, 2D garment contours are matched beforehand applying point registration based on Gaussian mixture models \cite{bing2010pointsetreg}. Finally the resulting mapped source image is sampled, and the segmented area can be superimposed using these colours. As a result, realistic rendering is provided showing both qualitative and quantitative advantages in relation to state-of-the-art method alternatives based on thin-plate splines with geodesic interpolation. The proposed retexturing method is visualised in Fig.~\ref{fig:flowChart}.

\begin{figure}
 \centering
 \begin{tabular}{ccc}
  {\includegraphics[width=.29\linewidth]{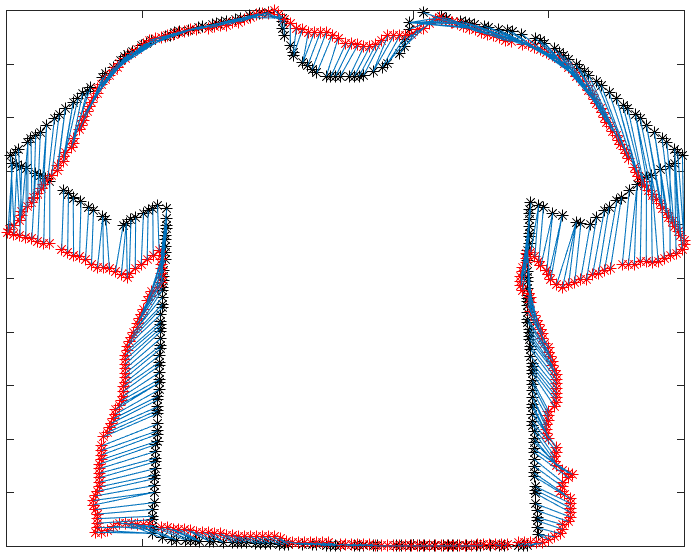}} &
  {\includegraphics[width=.29\linewidth]{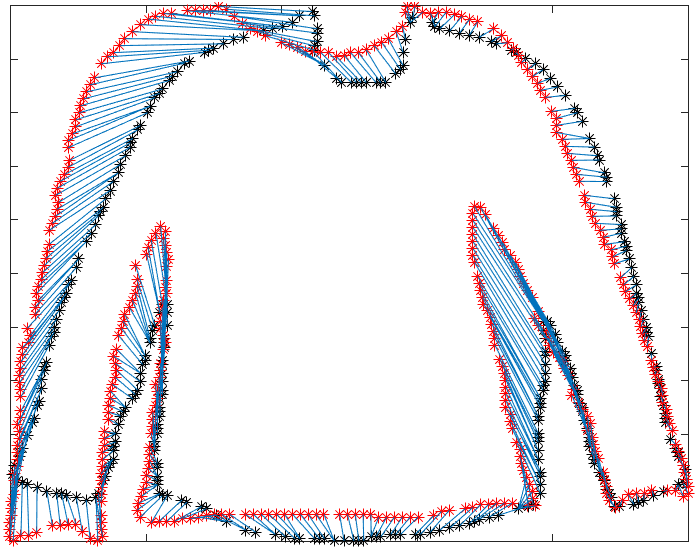}} &
  {\includegraphics[width=.29\linewidth]{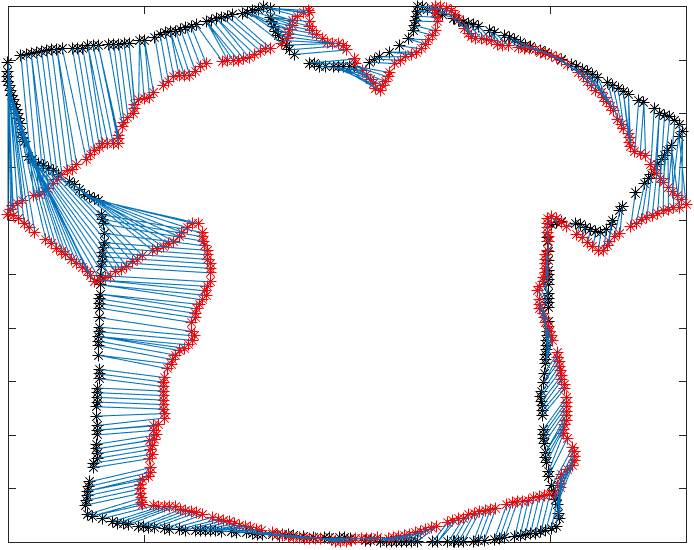}}  \\
  {\small{(a)}} &
  {\small{(b)}} &
  {\small{(c)}} 
 \end{tabular}
 \caption{Short and long sleeve examples for contour correspondences obtained using point set registration. Red contour corresponds to $C_R$ and blue contour corresponds to $C_F$.}
 \label{fig:contour22}
\end{figure}

\subsection{Segmentation}
In order to make accurate measurements in real world units, we standardise the coordinate system of body and garment models according to real world coordinates. Moreover, rich visualisation includes aligned image data (RGB and depth images), so as to provide animations as close as possible to the real scenario \cite{Henry2012}.

The first step of the proposed retexturing method is segmentation of garments from the background. It is necessary to extract a set of points from the image corresponding to the area being retextured. The proposed method works under the following assumptions: the area to be retextured is a shirt (or some other initially known garment) worn by a person, the person is assumed to be standing in front of the camera and is assumed not to occlude the area of interest with his/her hands. The segmentation is done by first extracting pixels and the skeleton of the body using Kinect SDK. Skeleton joint locations, along with some artificial joints, are used to train the GrabCut algorithm \cite{rother2004grabcut} and select areas with desired joints.
In the case of the reference 2D image, the GrabCut algorithm is also applied initializing the background color with the pixels on the borders of the image. The output of the grabCut algorithm is a binary mask, where the set of points are comprised of the outline of the binary mask. The initial point density is related to image resolution and area occupied by the garment. Typically the outline consists of few thousand points. This simple automatic segmentation approach worked accurately in our dataset. In case of other non-controlled scenarios, any other automatic or semi-automatic segmentation approach could be considered, such as deep learning based garment segmentation \cite{liang2015human} or pose guided garment boundaries. For cases where segmentation failed, we manually guided GrabCut to get the binary mask of the garment.

\begin{figure}[!ht]
  \centering
  \label{fig:geodesic}\includegraphics[width=0.95\linewidth]{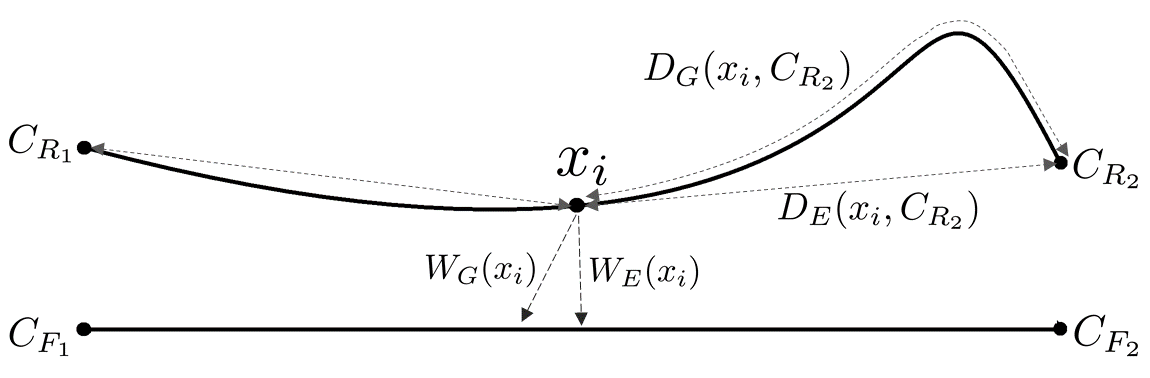}
  \caption{Comparing Euclidean to geodesic distace in TPS. TPS finds a mapping between two point sets based on known correspondences. In this image we consider such a mapping between two 2D example lines where end points $C_R$ are matched with $C_F$. As can be seen, point $x_i$ has equal Euclidean distance $D_E$ to points $C_{R_1}$ and $C_{R_2}$. In this case, mapping $W_E$ does not take line topology into account, causing a wrong interpolation where points on the right hand side of $W_E(x_i)$ get much denser than the points on its left hand side. This problem can be solved by using geodesic distance $D_G$ in the mapping $W_G$.}
\end{figure}

\subsection{Outer contour matching}
Contour matching can be viewed as a point set registration problem, where a correspondence must be found between a scene and a model. 
A few of the most well known methods for point set registration are iterative closest point~\cite{besl1992icp}, robust point matching~\cite{gold1998rpm, chui2003tpsrpm}, and Coherent point drift~\cite{myronenko2010} algorithms.
For our purposes, a correspondence must be found between highly deformed shapes. Out of available algorithms, we have chosen to use non-rigid point set registration using Gaussian mixture models (GMM) ~\cite{bing2010pointsetreg} because of its accurate fitting under different conditions and fast execution time. Additionally, Gaussian mixtures provide robust results even if the shapes have different features, such as different neck lines, hand positions and folds.

Let's define the contour of a garment on a real person as $C_R$ and the contour of the flat garment as $C_F$. The aim is to create a correspondence between contour models $C_R$ and $C_F$. 

In the GMM point matching algorithm, the point sets are represented as Gaussian mixture models. Instead of assuming a one-to-one correspondence based on the nearest neighbour criterion, one-to-many relaxations are used to allow for fuzzy correspondences, also known as soft assignment. The idea is to assume that each model point corresponds to a weighted sum of the scene points, instead of the closest scene point alone. The weights are proportional to a Gaussian function of the pairwise distances between the moving model and the fixed scene. The method works by dawning a statistical sample from a continuous probability distribution of random point locations. Afterwards the point set registration problem is viewed as an optimisation problem, meaning that a certain dissimilarity measure between the Gaussian mixtures constructed from the transformed model set and the fixed scene set is minimised based on L2 distance between the mixtures~\cite{bing2010pointsetreg}.

Before finding the corresponding points between the shapes, the contours point sets are reduced to 400 points. Afterwards, the point set x and y frame coordinates are normalised in the range [0,1]. Essentially the used method provides information about how $C_R$ has to be transformed to match $C_F$.  
Outer contour matching examples are shown in Fig. \ref{fig:contour22}.

\subsection{Inner contour matching}
Inner contour matching refers to the process of finding correspondence points between the body surface and the 2D flat garment in order to assign to each body point a colour from the garment. This process is mainly a difficult task due to, first, the lack of depth information for the 2D flat garment and the lack of texture for the depth image, and second, dissimilar textures for the source 2D flat garment and target put-on garment. Therefore feature based matching is not applicable. Conformal based approaches like \cite{Windheuser2011,zeng2008} fail due to the different topologies of the surfaces.

In order to solve this problem efficiently, we first generate a triangulated 3D mesh based on the depth image of the segmented area. To have a smooth shape at the boundaries of garment, we apply  morphological opening using disk structuring element type with mask size of 5 to the binary mask. A solution can be obtained by {finding an affine deformation matrix for each face triangle to bring both source and target surfaces into alignment} according to the matched points of the outer contours. {However, we cannot guarantee a perfect matching for near contour points in such a solution due to different surface topologies and depth camera noise in the contours. Instead, we propose to use} thin plate splines (TPS) \cite{bookstein1989} as a solution in closed-form based on a radial basis kernel. Let $X=\{x_1,...,x_N\}\in\mathbb{R}^3$ be the set of all points belonging to the segmented and discretized body surface $\Omega$. Then, a mapping from $x_i$ to the source image is computed through
\begin{equation}
W(x_i)=\sum_{j=1}^{n}\omega_j\kappa(\Arrowvert x_i-C_{R_j}\Arrowvert),
\end{equation}
where $\omega$ is a set of trained coefficients based on $C_R$ and $C_F$, $\kappa(d)=d^2\log d$ is a radial basis kernel and $n$ is the number of contour points. This basic formulation is based on Euclidean distance among the points which is not applicable for our problem since contour points do not cover all the surface; besides that, Euclidean distance does not describe the surface topology. Instead we propose a geodesic-based distance to include surface topology. We show this idea in Fig. \ref{fig:geodesic}. 

{Since we apply discretized body surface $\Omega$, the Dijkstra algorithm can be used to compute the shortest distance from $x_i$ to each $C_{R_j},j\in\{1..n\}$. However, we get a stairstep-like shortest path which introduces some amount of error in the distance, no matter how much we refine mesh. Instead, we follow the fast marching algorithm of \cite{Deschamps2001} to compute a fast and accurate approximation of geodesic distance. The fast marching algorithm is closely related to the Dijkstra algorithm with the difference that it satisfies the Eikonal equation $\Arrowvert\nabla U(x) \Arrowvert=1/s(x),x\in\Omega$ to update the graph where $\nabla U(x)$ is the gradient of the action map $U$ and $s(x)$ is a positive outwards speed function at point $x$. $U(x)$ is a function of time at point $x$ that describes the evolution of the surface with respect to $s(x)$ and surface gradient. We assume the surface is differentiable at all points. Starting from $x_i$, at each iteration, the algorithm sweeps outwards one grid point with respect to $s(x)$ to locate the proper grid point to update.} Then geodesic distance can be computed for two vertices $v_i$ and $v_j$ from the shortest path $L=\{L_1,...,L_m\}$ by
\begin{equation}
\Gamma(v_i,v_j)=\sum_{l=1}^{m-1}\Arrowvert L_l-L_{l+1}\Arrowvert
\end{equation}
To compute geodesic distance efficiently, we set a flag for cell $d_{ij}$ of the distance table as 1 if vertices $v_i$ and $v_j$ already exist on a larger optimum path, avoiding recomputing the optimum path for them.

Then we rewrite the TPS formulation to compute the coefficient matrix $\omega$ as
\begin{equation}\label{eq:tps}
\omega=\begin{bmatrix}\dot{K}_{n\times n}+\lambda I & [1|C_{R_{n\times 3}}]\\ [1|C_{R_{n\times 3}}]^\top& 0 \end{bmatrix}_{(n+4)^2}^{-1} \begin{bmatrix}C_{F_{n\times 3}}\\0 \end{bmatrix}_{(n+4)\times 3},
\end{equation}
where $\dot{K}_{ij}=\Gamma(C_{R_i},C_{R_j})^2\log\Gamma(C_{R_i},C_{R_j}) \forall i,j\in\{1,...,n\},i\neq j$. $\lambda I$ is a regularization term and is added to the kernel $\dot{K}$ where $I$ is the identity matrix and $\lambda\in\mathbb{R}$. $\lambda$ values close to zero make the kernel sensitive to wrong correspondences, and values far from zero tend to an affine transformation. We set $\lambda$ to -1000, and by doing so, the visualization becomes more realistic and less noisy. 

Afterwards, a solution can be achieved by applying trained coefficients as 
\begin{equation}
W=[\ddot{K}_{N\times n}|1|X_{N\times3}]\omega
\end{equation}
where $\ddot{K}_{ij}=\Gamma(X_i,C_{R_j})^2\log\Gamma(X_i,C_{R_j})$. Matrix $W$ includes warped points to the 2D shirt image. We assign each point the colour of its corresponding pixel from the shirt image.

\begin{figure*}[!ht]
  \centering
  \begin{subfigure}[]
{
	\label{fig:set1tshirts}\includegraphics[width=0.7\textwidth]{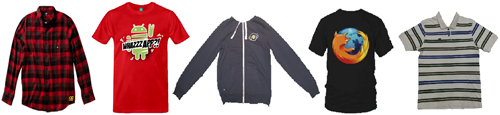}
}
\end{subfigure}
\begin{subfigure}[]
{
	\includegraphics[width=0.7\textwidth]{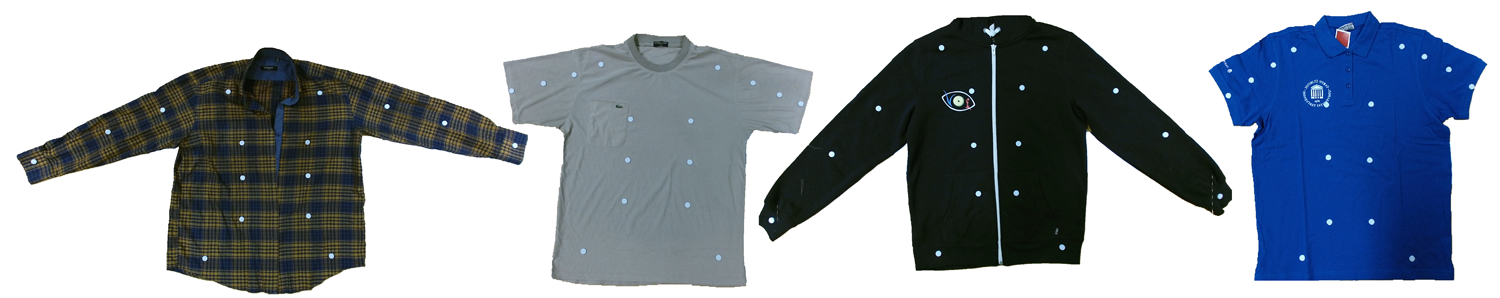}
  \label{fig:set1long}
}
\end{subfigure}
\begin{subfigure}[]
{
	\includegraphics[width=0.7\textwidth]{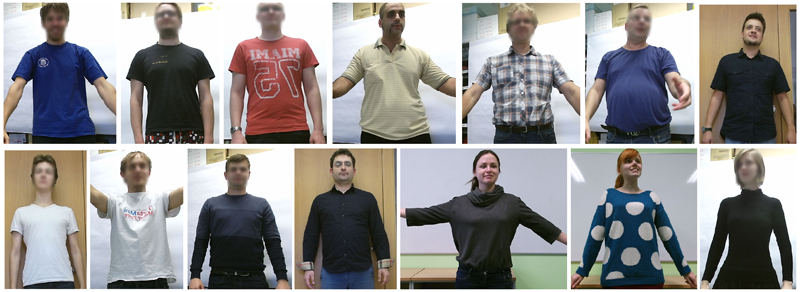}
  \label{fig:set1people}
}
\end{subfigure}
\caption{Sample images used in our dataset. a) Garments used in the first data set, b) Garments with landmarks used in the second dataset, c) People who participated in creating the first data set.}
\label{fig:db}
\end{figure*}

\subsection{Shading}
The shading effect of the garment is achieved using an adaptation of method \cite{egils2016autoretex} which is an automatic technique for garment retexturing and shading, where the shading information is acquired from Kinect~2 infrared information and is superimposed on the inner shape results. 
It is worth noticing that shadow mapping on the garment is not the main contribution of this paper, and thus its usage and coverage are limited to the extent
demanded for visualising the results illustrating the effectiveness of the proposed mapping method.

The general procedure for obtaining the final visualisation is as follows. The point cloud corresponding to the area of
interest provided by the Microsoft Kinect~2 camera is triangulated and rendered as described in the previous section.
The image created as a result of mapping in the previous steps is used as a texture image, such that each vertex corresponds to a point
on the image. Afterwards, the rendered image is modified by the corresponding infrared values for each pixel. Finally, the segmented area in the Kinect frame is replaced by the colour information from the previous step.

In order to enhance the quality of the representation, the point cloud is preprocessed before rendering, since it usually is noisy. More clearly, smoothing the depth image with a Gaussian filter is considered, which, according to our experiments, significantly improves the results.

\section{Experimental Results and Discussion}

In order to present the results, first we describe the setup of the experiments in terms of data, methods and parameters and evaluation metrics.

\subsection{Setup}

The proposed retexturing method was tested on an image database taken using the Kinect~2 RGB-D camera. According to~\cite{yang2015kinect}, Kinect 2 can capture frames starting from 0.5 meters and has depth accuracy error smaller than 2 mm in the cater part of the frame. The error increases towards edges of the frame, and it also increases with greater measurement distances. The best distance for scanning objects is the 0.5 to 2m range. To achieve the best depth resolution, the people were scanned at a distance of 1.5 to 2 meters where the error in the horizontal and vertical plane is the smallest. Each image contains a person facing the camera in a pose that does not significantly occlude the worn garment. The garments segmented from the original database were retextured using another database consisting of images of flat shirts. The flat shirt database was captured with various cameras providing decent quality images, as depth was not required.

\begin{figure*}[!ht]
  \centering
  \begin{subfigure}[]
{
	\includegraphics[width=0.45\textwidth]{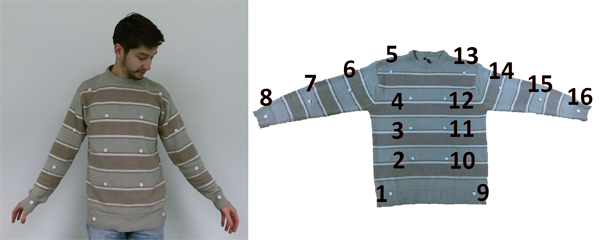}
  \label{fig:lmarks1}
}
\end{subfigure}
\begin{subfigure}[]
{
	\includegraphics[width=0.45\textwidth]{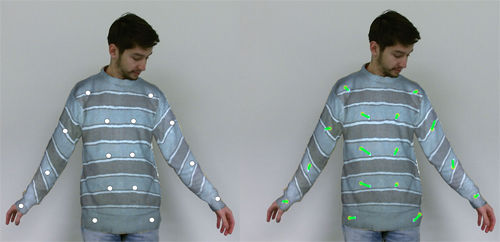}
  \label{fig:lmarks2}
}
\end{subfigure}
  \caption{a) Landmarks used in the second dataset on the flat garment (right) and landmark locations after putting the garment on as ground truth. Landmarks are shown by indices for comparison purposes. b) Retextured garment and estimated landmarks (left) and displacement arrows to ground truth landmarks (right) for computing error.}
  \label{fig:lmarks}
\end{figure*}

\begin{table*}[!ht]
 \caption{Mean Opinion Score (MOS) comparison}
 \label{fig:MOS}
 \centering

 \begin{tabular}{ | c | c | c | c | c | }
    \hline
    Method & T-shirt Votes & T-shirt Percentage & Long sleeve Votes & Long sleeve Percentage \\ \hline
    NRICP \cite{amberg2007} & 77 & 2.68\% & 32 & 3.69\% \\ \hline
    CPD \cite{myronenko2010} & 485 & 16.88\% & 245 & 28.23\% \\ \hline
    GM-TPS & 2311 & 80.44\% & 591 & 68.09\% \\ 
    \hline
  \end{tabular}
\end{table*}

The first data set contained 91 retextured images with 14 people (11 males and 3 females). This data set used 13 flat garments (4 long sleeve garments and 9 t-shirts). The second data set contained 39 retextured images with 5 people (4 males and 1 female). This data set used 8 flat garments (4 long sleeve garments and 4 t-shirts). We physically attached 16 landmarks to garments in the second dataset. The location of the landmarks was chosen empirically with the aim of visually demonstrating the texture shifts for different parts of the garment. Matching the landmarks was a manual process, therefore we limited ourselves to 16 landmarks. This comparison was done in order to determine the retexturing precision by retexturing the same garment onto itself. In ideal case the retextured image should be identical to the original image. Fig. \ref{fig:lmarks1} shows a sample of a real put-on image and the landmarked garment itself. Some samples of both datasets are shown in Fig. \ref{fig:db}.

\begin{figure*}
 \centering
 \begin{tabular}{ m{.15\textwidth} m{.15\textwidth} m{.15\textwidth} m{.15\textwidth} m{.15\textwidth} m{.15\textwidth} m{.01\textwidth} }
  {\includegraphics[width=.1\textwidth]{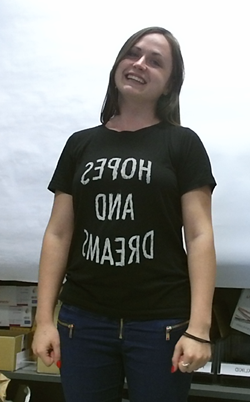}} &
  {\includegraphics[width=.1\textwidth]{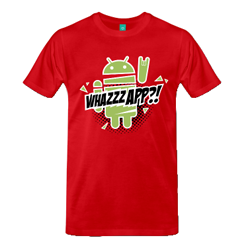}} &
  {\includegraphics[width=.1\textwidth]{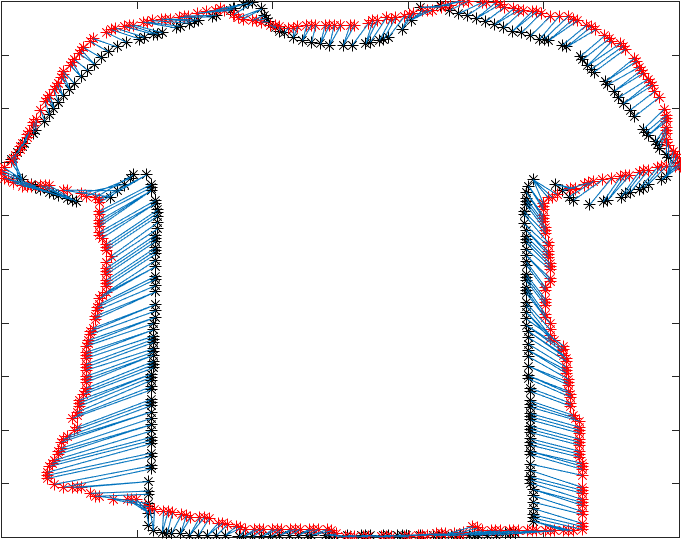}} &
  {\includegraphics[width=.1\textwidth]{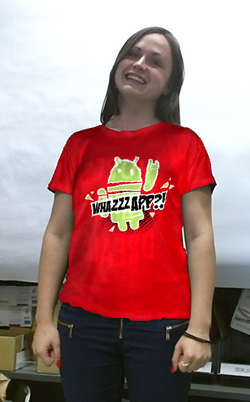}} &
  {\includegraphics[width=.1\textwidth]{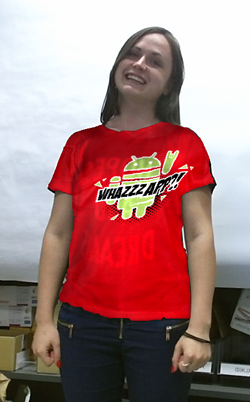}} &
  {\includegraphics[width=.1\textwidth]{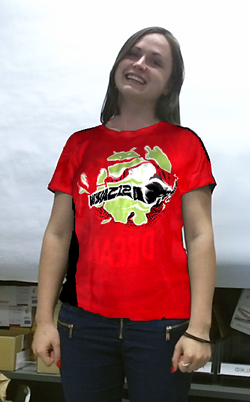}} \\
  
  {\includegraphics[width=.1\textwidth]{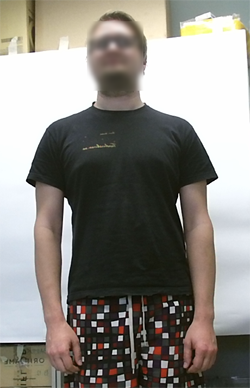}} &
  {\includegraphics[width=.1\textwidth]{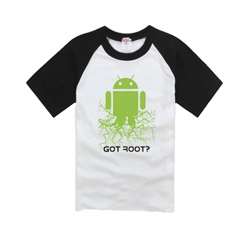}} &
  {\includegraphics[width=.1\textwidth]{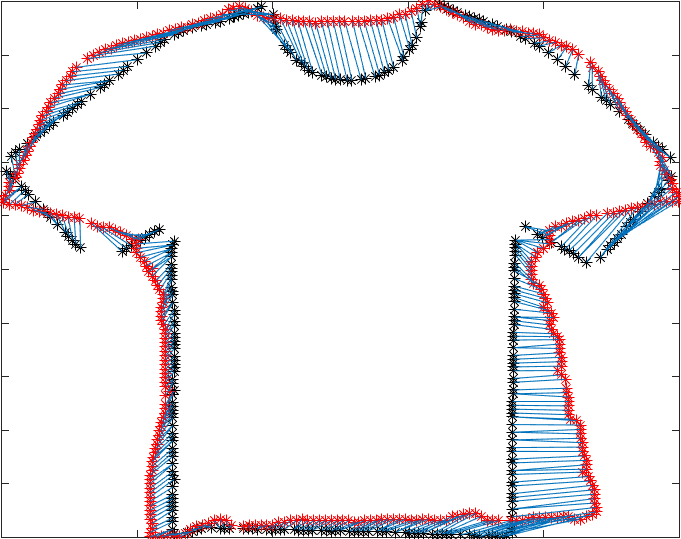}} &
  {\includegraphics[width=.1\textwidth]{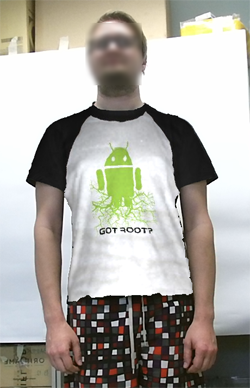}} &
  {\includegraphics[width=.1\textwidth]{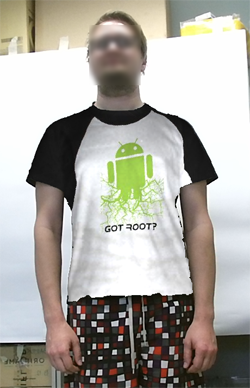}} &
  {\includegraphics[width=.1\textwidth]{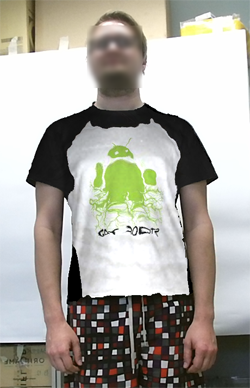}} \\
  
  {\includegraphics[width=.1\textwidth]{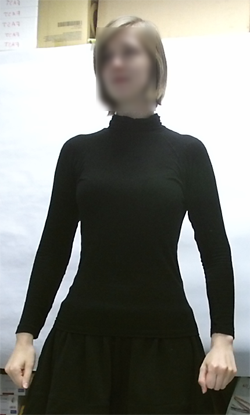}} &
  {\includegraphics[width=.1\textwidth]{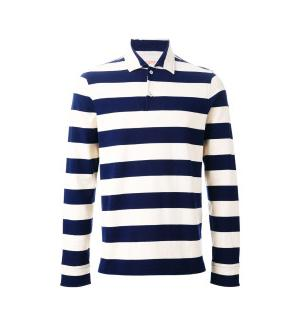}} &
  {\includegraphics[width=.1\textwidth]{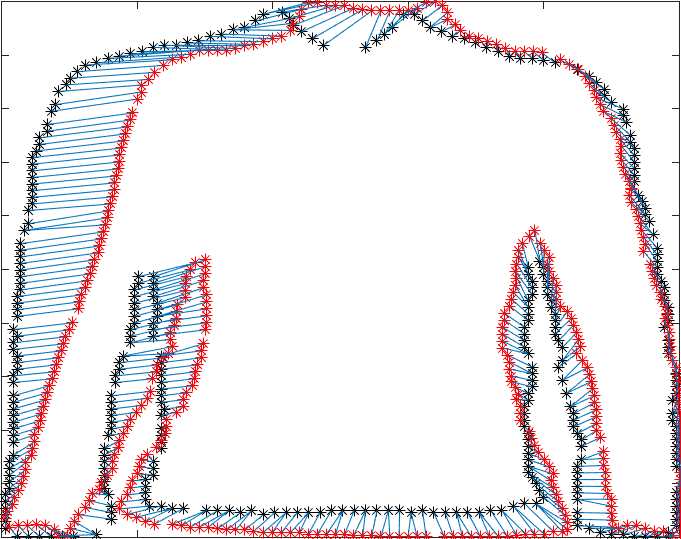}} &
  {\includegraphics[width=.1\textwidth]{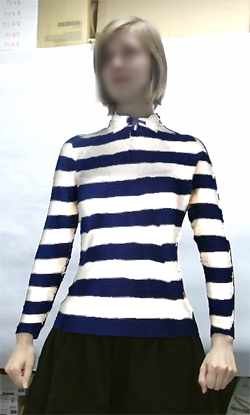}} &
  {\includegraphics[width=.1\textwidth]{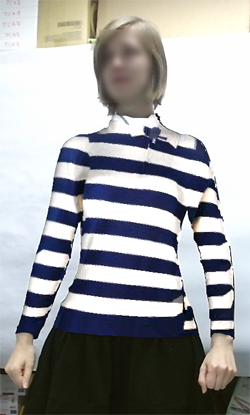}} &
  {\includegraphics[width=.1\textwidth]{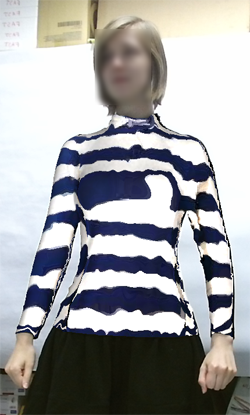}} \\

  {\includegraphics[width=.1\textwidth]{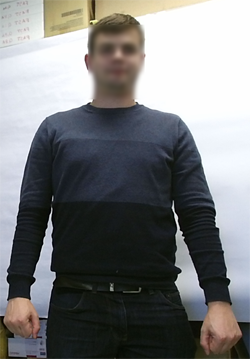}} &
  {\includegraphics[width=.1\textwidth]{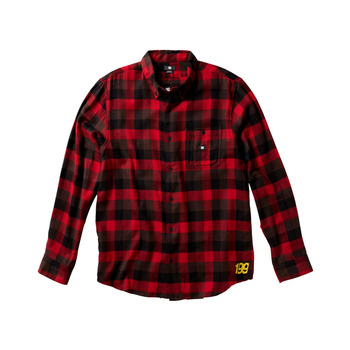}} &
  {\includegraphics[width=.1\textwidth]{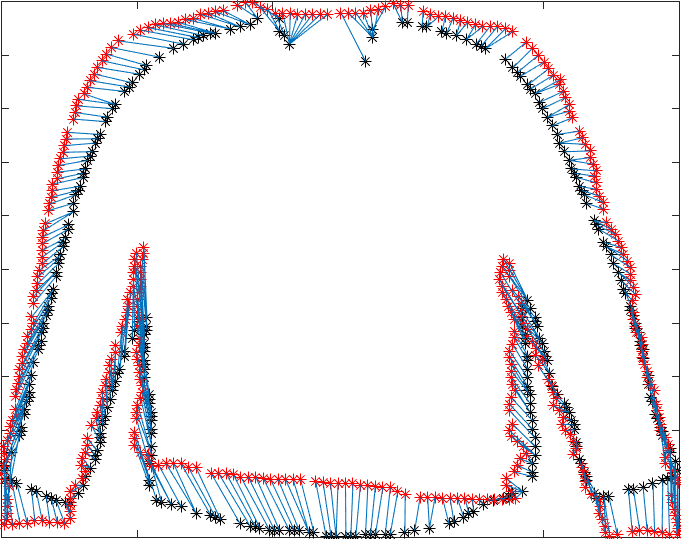}} &
  {\includegraphics[width=.1\textwidth]{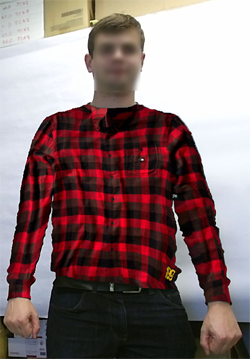}} &
  {\includegraphics[width=.1\textwidth]{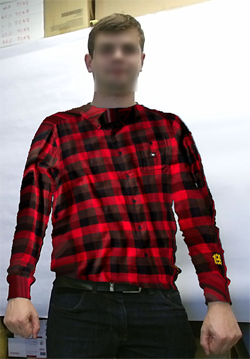}} &
  {\includegraphics[width=.1\textwidth]{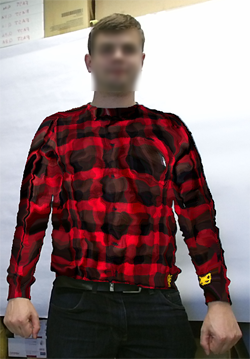}} \\

  {\includegraphics[width=.1\textwidth]{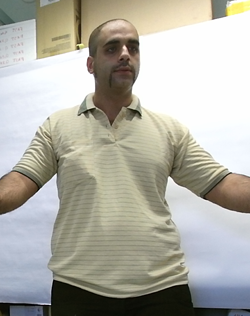}} &
  {\includegraphics[width=.1\textwidth]{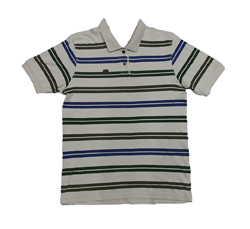}} &
  {\includegraphics[width=.1\textwidth]{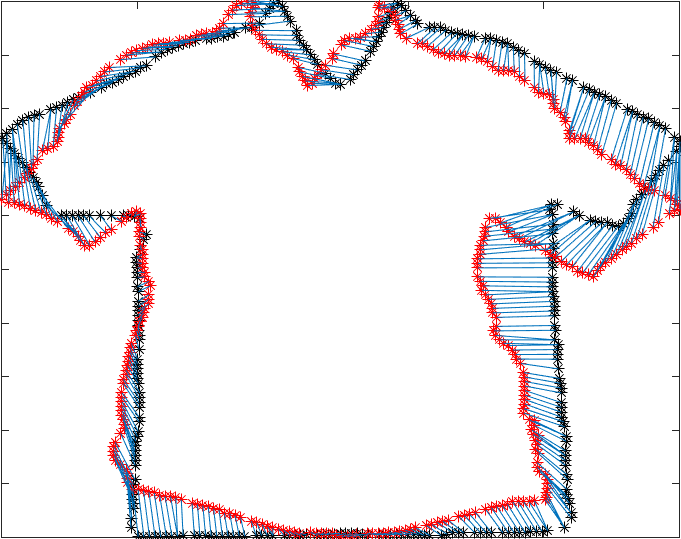}} &
  {\includegraphics[width=.1\textwidth]{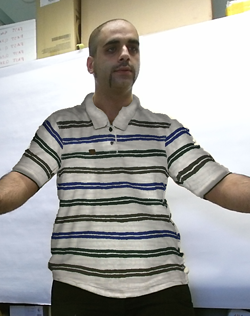}} &
  {\includegraphics[width=.1\textwidth]{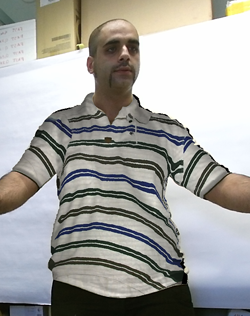}} &
  {\includegraphics[width=.1\textwidth]{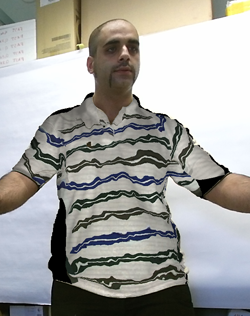}} \\

  {\includegraphics[width=.1\textwidth]{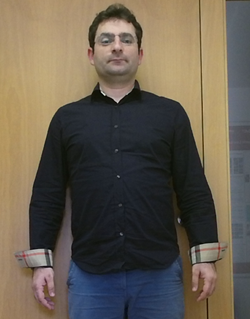}} &
  {\includegraphics[width=.1\textwidth]{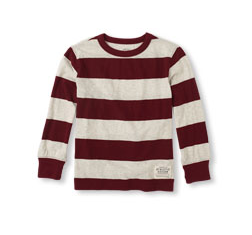}} &
  {\includegraphics[width=.1\textwidth]{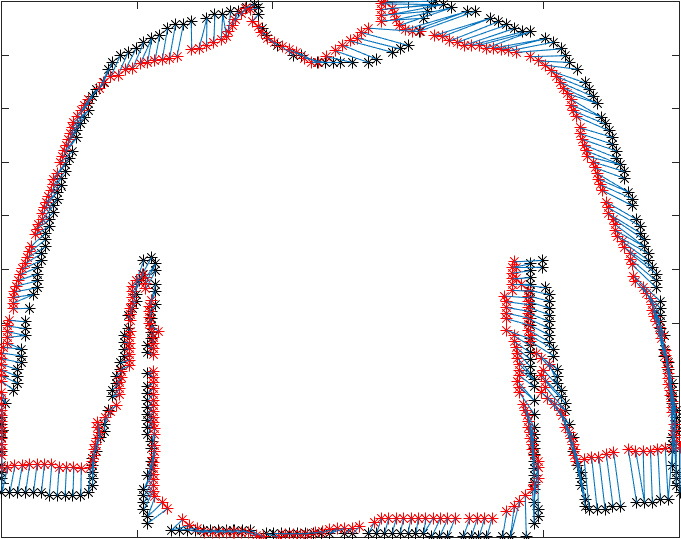}} &
  {\includegraphics[width=.1\textwidth]{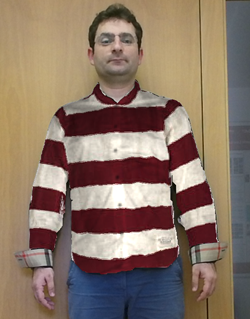}} &
  {\includegraphics[width=.1\textwidth]{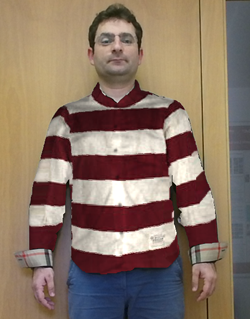}} &
  {\includegraphics[width=.1\textwidth]{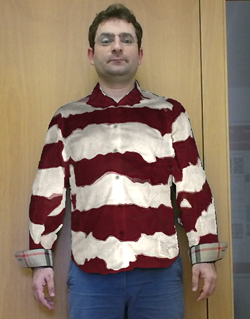}} \\
  {\includegraphics[width=.1\textwidth]{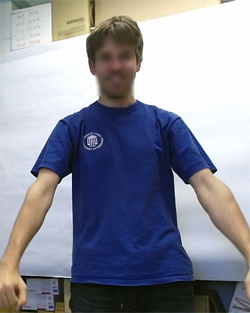}} &
  {\includegraphics[width=.1\textwidth]{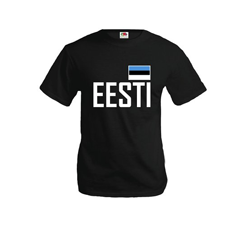}} &
  {\includegraphics[width=.1\textwidth]{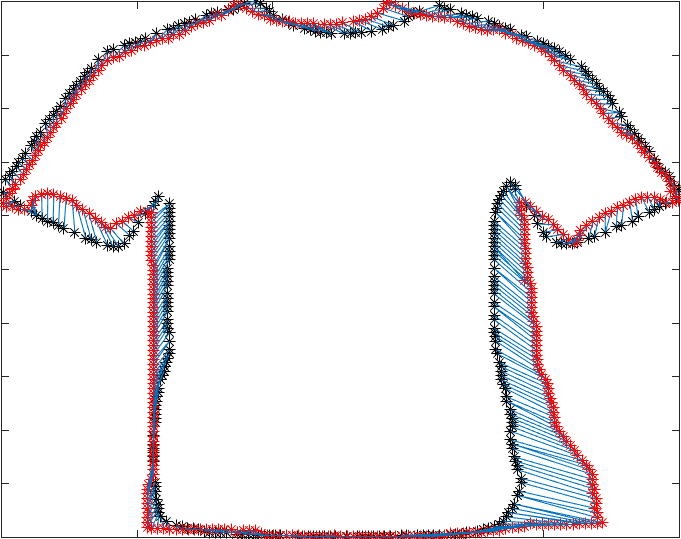}} &
  {\includegraphics[width=.1\textwidth]{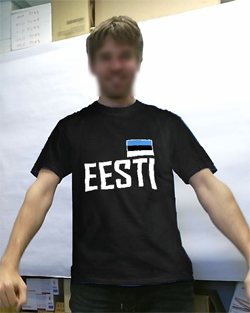}} &
  {\includegraphics[width=.1\textwidth]{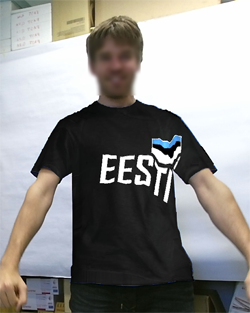}} &
  {\includegraphics[width=.1\textwidth]{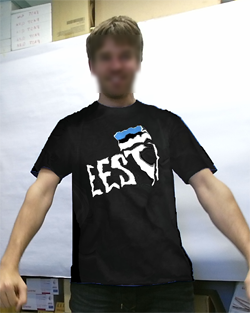}} \\
  \centering {\small{(a)}} &
  \centering {\small{(b)}} &
  \centering {\small{(c)}} &
  \centering {\small{(d)}} &
  \centering {\small{(e)}} &
  \centering {\small{(f)}} & \\
 \end{tabular}
 \caption{Images created by the proposed retexturing method, (a) is the original image, (b) is the image of a shirt, (c) shows the shape correspondence, (d) is the retextured image based on the geodesic mapping, (e) is mapping using the Coherent Point Drift (CPD) algorithm and (f) is mapping using the non-rigid Iterative Closest Point (ICP) algorithm.}
 \label{fig:results}
\end{figure*}

We used two metrics for evaluation of our method: qualitative comparison using the mean opinion score (MOS), and quantitative comparison using the mean square error (MSE). The MOS score was measured by showing 91 sets of images from the first dataset to 41 people. The data was presented in an online survey where the image size was two times larger to that of shown in \ref{fig:results}, with the exception of column (c) which was not shown to the participants. Also it has to be pointed out, that most of the participants did not have educational background in image processing or related fields. In the survey each person was asked for an opinion about which one of the images in each set looks visually more realistic. The MSE was measured on the second dataset by retexturing the flat version of the shirt and computing the average distance from retextured landmarks to ground truth landmarks. Fig. \ref{fig:lmarks2} shows the process of computing MSE. Unfortunately, some retexturing or garment fitting papers just report results as a few qualitative images \cite{zhou2012image, hauswiesner2011free, Zhang2013}. Regarding our contribution as a garment point matching, we select point set registration methods in the comparison using introduced evaluation metrics: nonrigid iterative closest point (NRICP) \cite{amberg2007} and coherent point drift (CPD) \cite{myronenko2010}.

All compared results were produced with the same set of parameters that were determined empirically.
The setup parameters for matching the contours needed for the point registration algorithm \cite{bing2010pointsetreg} are set as follows {(see original paper for the definition of parameters)}: sigma, which is the scale parameter of Gaussian mixtures, is set to 0.2 and 0.1, and the maximum number of function evaluations at each level is set to 50, 500, 100, 100 and 100. The point registration algorithm uses contours with 400 points. After the transformation and point correspondence are found, the contour is further down-sampled to 120 points and used for the inner point matching. A larger number would have resulted
in a long computation time, whereas a smaller number of points resulted in some undersampled parts and produced inferior mappings. 120 points were chosen as a compromise between the execution time and the resulting mapping quality.

\setlength\tabcolsep{3pt}

\setlength\tabcolsep{3pt}
\begin{figure*}
 \centering
 \begin{tabular}{ m{0.18\textwidth} m{0.18\textwidth} m{0.18\textwidth} m{0.15\textwidth}}
 
 
  
  {\includegraphics[width=0.16\textwidth]{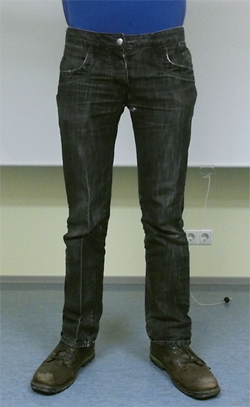}} &
  {\includegraphics[width=0.16\textwidth]{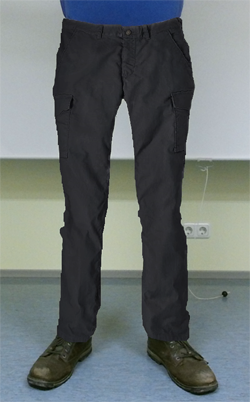}} &
  {\includegraphics[width=0.18\textwidth]{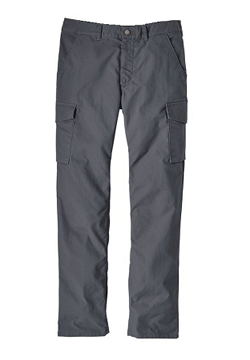}} &
  {\includegraphics[width=0.15\textwidth]{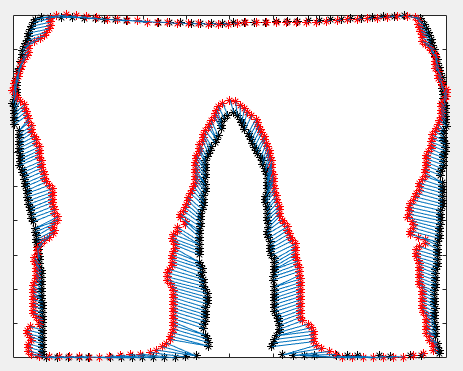}} \\
  
  {\includegraphics[width=0.16\textwidth]{images/pants/eric1o}} &
  {\includegraphics[width=0.16\textwidth]{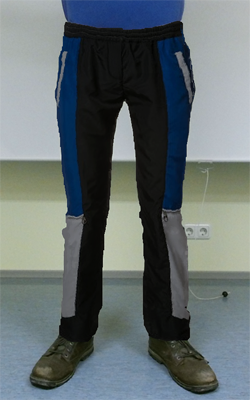}} &
  {\includegraphics[width=0.18\textwidth]{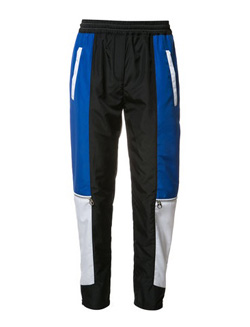}} &
  {\includegraphics[width=0.15\textwidth]{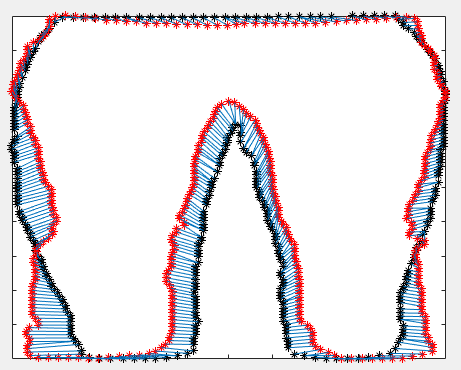}} \\
  
  
  {\includegraphics[width=0.18\textwidth]{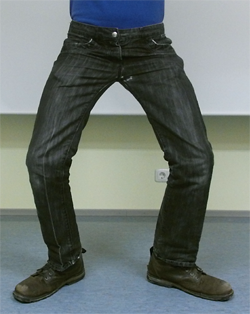}} &
  {\includegraphics[width=0.18\textwidth]{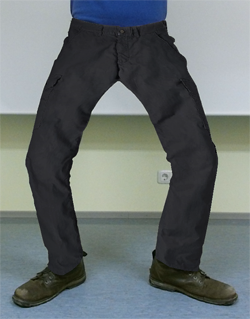}} &
  {\includegraphics[width=0.18\textwidth]{images/pants/pants3}} &
  {\includegraphics[width=0.15\textwidth]{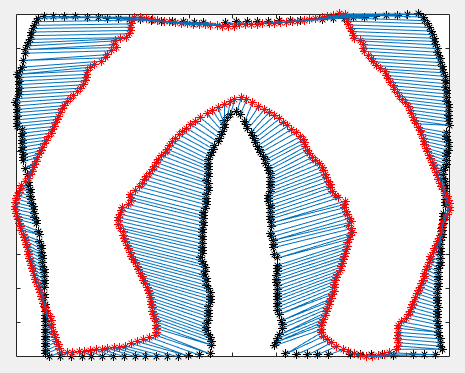}} \\
  
  {\includegraphics[width=0.18\textwidth]{images/pants/eric3o}} &
  {\includegraphics[width=0.18\textwidth]{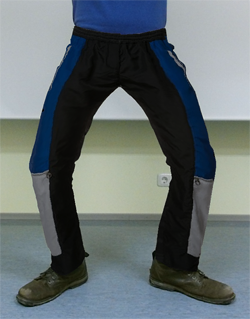}} &
  {\includegraphics[width=0.18\textwidth]{images/pants/pants4}} &
  {\includegraphics[width=0.15\textwidth]{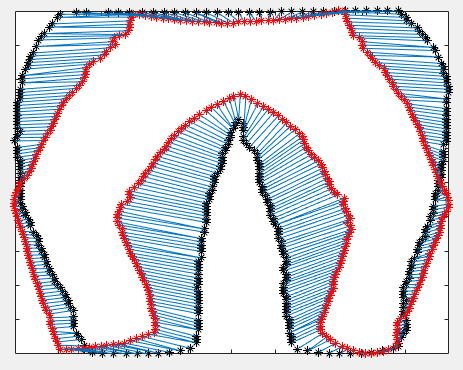}} \\
  
  \centering {\small{(a)}} &
  \centering {\small{(b)}} &
  \centering {\small{(c)}} &
  \centering {\small{(d)}} \\
 \end{tabular}
 \caption{Images created by the proposed retexturing method, (a) is the original image, (b)  is the retextured image based on the geodesic mapping, (c) is the image of pants and (d) shows the shape correspondence.}
 \label{fig:pants}
\end{figure*}

\begin{figure*}[!t]
  \centering
  \includegraphics[width=0.9\textwidth]{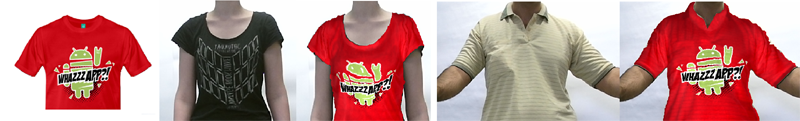}
  \caption{Retexturing effects for different necklines}
  \label{fig:necklines}
\end{figure*}

\subsection{Evaluation}
We separated long and short sleeve images in the results to analyse them separately. We show the MOS percentage in Table \ref{fig:MOS}. The results illustrate that our method outperforms state-of-the-art methods by a large margin regarding realistic view. This can also be seen qualitatively in Fig. \ref{fig:results}. We added correspondences between flat garment and body contours in the third column of Fig. \ref{fig:results} to see the effect of outer contour matching on the retexturing results. It can be seen that the final retextured image still has a realistic appearance even with small misalignment in outer correspondences. However, a small misalignment can have a local impact. This can mainly be seen in the long sleeves. As an additional qualitative example, Fig. \ref{fig:pants} shows the proposed retexturing approach used to successfully retexture pants. Given the appropriate input data, the same can be done for other garment types. It has to be noted that the appearance of shadows are highly dependant on the material of the garment a person is wearing. The adjustment of Kinect IR values uses the same parameters for all generated results, therefore the shadow effects can appear different for different garment types. The use of Kinect IR data for shadow generations is the same as the one presented in~\cite{traumann2015new, egils2016autoretex}.

If the source and target garments have different features, for example if a collar is present in the put-on image and not present in the flat image, some unnatural effects may be seen; the same goes for different neck lines as shown in Fig \ref{fig:necklines}. The NRICP algorithm has the worst visual results due to the different topologies of the surfaces between flat garment and body, and the CPD algorithm has difficulties with aligning surfaces in the boundary regions. 

\begin{figure*}[!ht]
  \centering
  \begin{subfigure}[]
{
	\includegraphics[width=.45\textwidth]{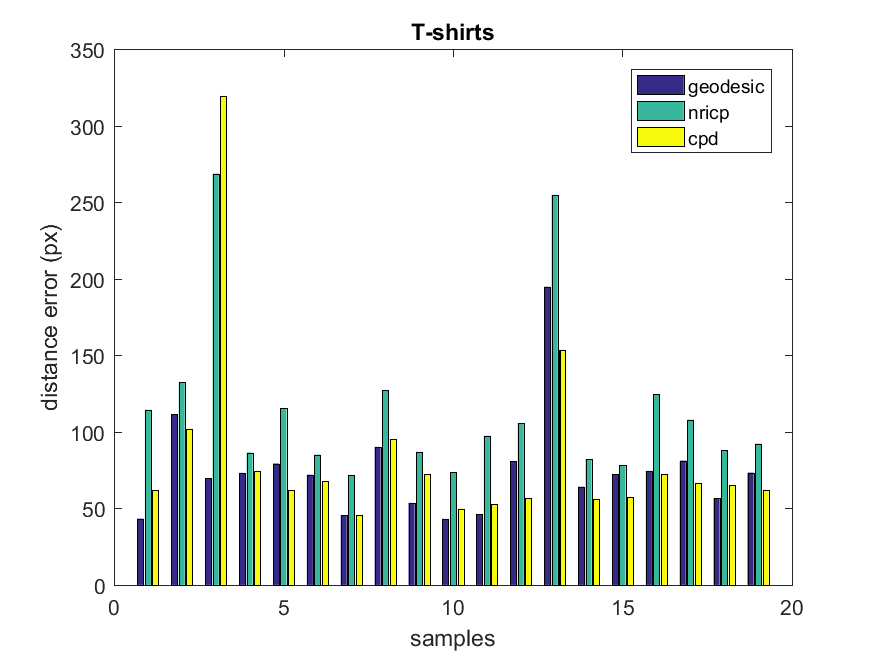}
  \label{fig:t-dist}
}
\end{subfigure}
\begin{subfigure}[]
{
	\includegraphics[width=.45\textwidth]{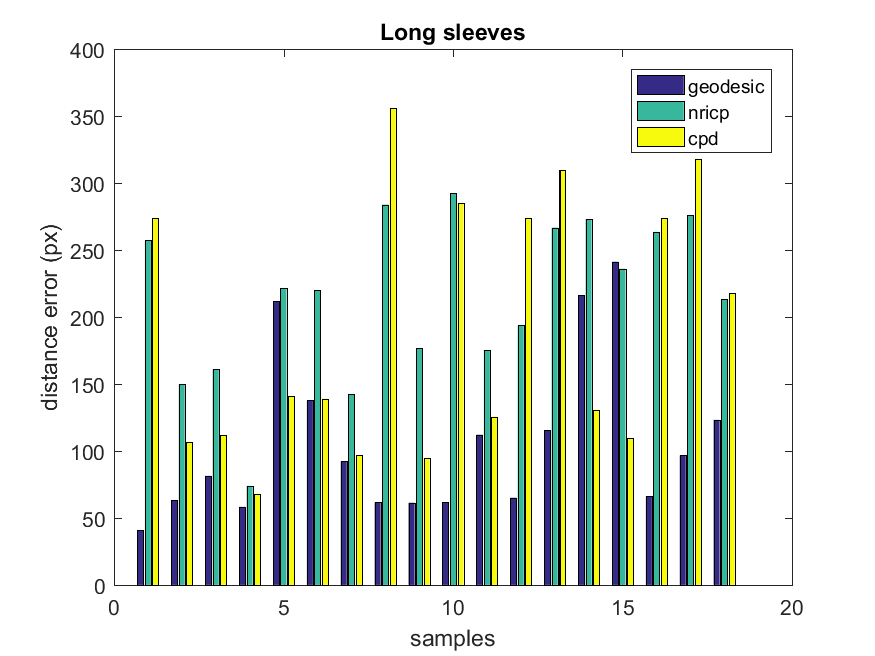}
  \label{fig:l-dist}
}
\end{subfigure}
\begin{subfigure}[]
{
	\includegraphics[width=.45\textwidth]{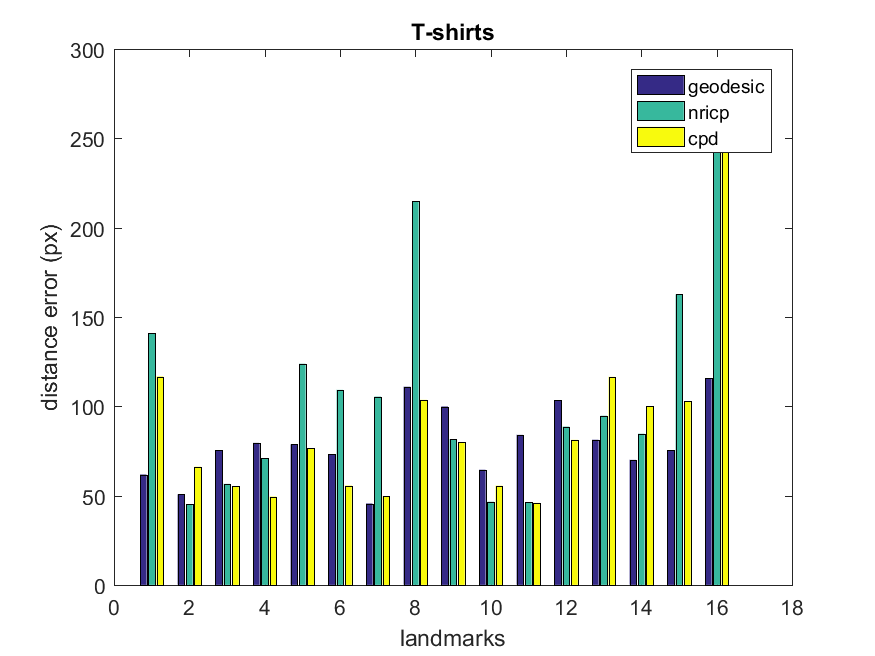}
  \label{fig:t-landmark}
}
\end{subfigure}
\begin{subfigure}[]
{
	\includegraphics[width=.45\textwidth]{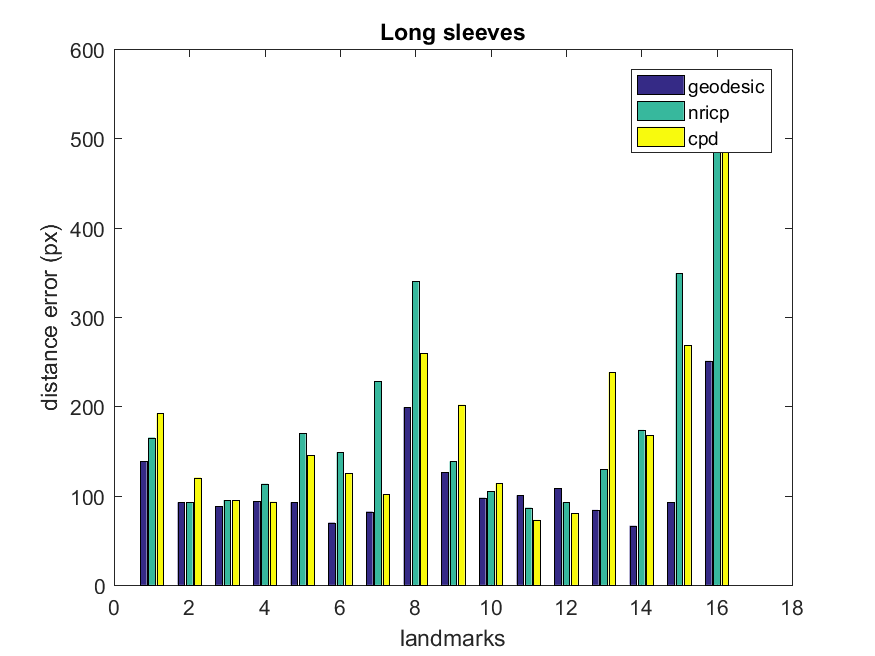}
  \label{fig:l-landmark}
}
\end{subfigure}
  \caption{Method comparison for different garments. Graph (a) and (b) show average landmark error (in pixels) for each sample for T and long sleeve shirts. Graphs (c) and (d) show per landmark error averaged over all samples for T and long sleeve shirts.}
  \label{fig:contour}
\end{figure*}

MSE values are shown in Table \ref{fig:err}. As seen from the visual results, in most cases our method is more accurate than state-of-the-art methods regarding marker distances to ground truth. Our method generates a lower error for short sleeves than long sleeves.
However, this is not a significant change according to the MSE results. Often our method performs better than other methods for almost each marker in Fig. \ref{fig:contour} where samples represent different garments, and landmarks are the white circles that are placed on the garment, as is shown in Fig. \ref{fig:lmarks}. 
Our method is more stable among different persons and different markers in comparison to the state-of-the-art methods. However, the long sleeves error as seen in Fig.~\ref{fig:l-dist} fluctuates among different persons due to higher variation in hand position. Marker numbers 8 and 16, which were placed at the end of the sleeves, have the highest error in both long and short sleeve garments. This happens due to slight point misalignment in outer contour matching.

\begin{table}
 \caption{MSE for marker mapping error on the second dataset}
 \label{fig:err}
 \centering

 \begin{tabular}{ | c | c | c | }
    \hline
    Method & T-shirts  & Long sleeves\\ \hline
    NRICP \cite{amberg2007} & $115.400$ px  & $215.349$ px\\ \hline
    CPD \cite{myronenko2010} & $83.850$ px  & $190.618$ px\\ \hline
    GM-TPS & $75.005$ px  & $105.884$ px\\
    \hline
  \end{tabular}
\end{table}

\subsection{Time complexity}
We analyze time complexity of the retexturing part, the main contribution of this work. Computing geodesic distance is the most time consuming part of retexturing. The basic fast marching algorithm has a time complexity $O(N\log(N))$ for Eikonal solver where $N$ is the number of nodes in the mesh. Yatziv et al. \cite{yatziv2006n} reduced the complexity to $O(N)$ via untidy priority queue. Although basic formulation does not allow parallel computing, near optimal iterative Eikonal solvers have been appeared for fast and parallel computing \cite{hong2016multi}. Fu et al. \cite{fu2011fast} reported a computation time 459ms for a Stanford dragon with 631,187 vertices speeding up basic fast marching algorithm by a factor of 14. Note that a garment in our setup has 15K vertices in average. Without loss of generality one can resize depth image by a factor of 0.5 and reduce number of vertices less than 4K, meaning a geodesic computation in less than 3ms for a single node. However, we need to compute geodesic distance for all $N$ vertices which makes it polynomial time complexity. Fortunately having a table of shortest distances among all $N$ vertices which is getting updated iteratively allows us to reduce polynomial time complexity to $N\log(N)$, assuring real time computation performance.

\section{Conclusion}
We proposed a retexturing method based on robust point registration and thin plate spline interpolation. 
The proposed method can be used to segment out the garment worn by a person and retexture it with another similar piece of garment, i.e. a garment lying on the table or some other flat surface. In this fashion, the outer boundaries of the segmented put-on garment are matched to the boundaries of the flat source garment. Afterwards, the whole surfaces are matched based on geodesic thin plate spline to assign each point on the target garment a color from the source garment. We compared our approach to the state-of-the-art methods and achieved the best results in both visual and numerical evaluations. 

Our current approach is limited to a relaxed pose without occlusions on the garment. However, our approach is general as long as boundary correspondences are given. In future work, we will consider a 3D human model fitting to cope with current limitations. 

\section*{Acknowledgment}
This work has been partially supported by Estonian Research Council Grant PUT638, Fits.Me (Rakutan) through the Research and Development Project LLTTI16056, the Spanish Projects TIN2015-65464-R and TIN2016-74946-P (MINECO/FEDER, UE), CERCA Programme / Generalitat de Catalunya, the Scientific and Technological Research Council of Turkey (TÜBİTAK) 1001 Project (116E097), the COST Action IC1307 iV\&L Net (European Network on Integrating Vision and Language) supported by COST (European Cooperation in Science and Technology), and the Estonian Centre of Excellence in IT (EXCITE) funded by the European Regional Development Fund.

\bibliographystyle{ieee}
\bibliography{refs}
\end{document}